\documentclass[10pt,twocolumn,letterpaper]{article}

\usepackage{cvpr}
\usepackage{times}
\usepackage{epsfig}
\usepackage{graphicx}
\usepackage{amsmath}
\usepackage{amssymb}
\usepackage{xcolor}
\usepackage{soul}
\usepackage{listings}

\usepackage[pagebackref=true,breaklinks=true,letterpaper=true,colorlinks,bookmarks=false]{hyperref}

\cvprfinalcopy 

\def\cvprPaperID{5526} 

\newcommand{\figref}[1]{Figure~\ref{fig:#1}}
\newcommand{\tabref}[1]{Table~\ref{tab:#1}}
\newcommand{\eqnref}[1]{Equation~\ref{eqn:#1}}
\newcommand{\secref}[1]{Section~\ref{sec:#1}}

\newcommand\Tstrut{\rule{0pt}{2.4ex}}       

\newcommand{\ADD}[1]{#1}

\ifcvprfinal\fi

\definecolor{codegreen}{rgb}{0,0.6,0}
\definecolor{codegray}{rgb}{0.5,0.5,0.5}
\definecolor{codepurple}{rgb}{0.58,0,0.82}
\definecolor{backcolour}{rgb}{0.95,0.95,0.92}

\lstdefinestyle{mystyle}{
	backgroundcolor=\color{backcolour},   
	commentstyle=\color{codegreen},
	keywordstyle=\color{magenta},
	numberstyle=\tiny\color{codegray},
	stringstyle=\color{codepurple},
	basicstyle=\footnotesize,
	breakatwhitespace=false,         
	breaklines=true,                 
	captionpos=b,                    
	keepspaces=true,                 
	numbers=left,                    
	numbersep=5pt,                  
	showspaces=false,                
	showstringspaces=false,
	showtabs=false,                  
	tabsize=2
}

\begin{document}
	
	\title{Single-frame Regularization for Temporally Stable CNNs}
	
	
	\author{Gabriel Eilertsen$^1$ \hspace{1.2cm} Rafa\l~K. Mantiuk$^2$ \hspace{1.2cm} Jonas Unger$^1$\\
		$^1$Dept. of Science and Technology, Link\"oping University, Sweden\\
		$^2$Dept. of Computer Science and Technology, University of Cambridge, UK\\
		{\tt\small \{gabriel.eilertsen, jonas.unger\}@liu.se \hspace{0.1cm} rafal.mantiuk@cl.cam.ac.uk}
	}
	
	\maketitle
	
	\begin{abstract}
		Convolutional neural networks (CNNs) can model complicated non-linear relations between images. However, they are notoriously sensitive to small changes in the input. Most CNNs trained to describe image-to-image mappings generate temporally unstable results when applied to video sequences, leading to flickering artifacts and other inconsistencies over time. In order to use CNNs for video material, previous methods have relied on estimating dense frame-to-frame motion information (optical flow) in the training and/or the inference phase, or by exploring recurrent learning structures. 
		We take a different approach to the problem, posing temporal stability as a regularization of the cost function. The regularization is formulated to account for different types of motion that can occur between frames, so that temporally stable CNNs can be trained without the need for video material or expensive motion estimation. The training can be performed as a fine-tuning operation, without architectural modifications of the CNN. Our evaluation shows that the training strategy leads to large improvements in temporal smoothness.
		Moreover, for small datasets the regularization can help in boosting the generalization performance to a much larger extent than what is possible with na\"ive augmentation strategies.
	\end{abstract}
	
	\section{Introduction}
	Deep neural networks (DNNs) can represent complex non-linear functions but tend to be very sensitive to the input. For image data, 
	this is manifested in sensitivity to small changes in pixel values. For example, techniques for generating adversarial examples have demonstrated that there exists images that are visually indistinguishable from each other, while generating widely different predictions \cite{Szegedy2013}. It is also possible to find naturally occurring image operations that can cause a convolutional neural network (CNN) to fail in the learned task \cite{Azulay2018,Engstrom2017,Zheng2016}. For image-to-image CNNs applied on video sequences, this sensitivity results in abrupt and incoherent changes from frame to frame. Such changes are seen as flickering, or unnatural movements of local features. 
	Previous methods for applying CNNs to video material most often use dense motion information between frames in order to enforce temporal coherence \cite{Ruder2016,Gupta2017,Chen2017,Lai2018}. This requires ground truth optical flow for training, modifications to the CNN architecture, and computationally expensive training and/or prediction. Moreover, there are many situations where reliable correspondences between frames cannot be estimated, \eg due to occlusion or lack of texture.
	
	Instead of relying on custom architectures, we take a simple, efficient, and general approach to the problem of CNN temporal stability. We pose the stability as a regularizing term in the loss function, which potentially can be applied to any CNN. We formulate two different regularizations based on observations of the expected behavior of temporally varying processing. 
	The result is a light-weight method for stabilizing CNNs in the temporal domain. It can be applied through fine-tuning of pre-trained CNN weights and requires no special-purpose training data or CNN architecture. Through extensive experimentation for the application in colorization and single-exposure high dynamic range (HDR) reconstruction, we show the efficiency of the regularization strategies. 
	
	In summary, this paper explores regularization for the purpose of stabilizing CNNs in the temporal domain and presents the following main contributions:
	
	\begin{itemize}
		\item Two novel regularization formulations for temporal stabilization of CNNs, which both model the dynamics of consecutive frames in video sequences.
		\item A novel perceptually motivated smoothness metric for evaluation of the temporal stability.
		\item An evaluation showing that the proposed training technique improves temporal stability significantly while maintaining or even increasing the CNN performance.
		\item For scenarios with limited training data, the generalization performance of the regularization strategies is significantly better than traditional data augmentation.
	\end{itemize}
	
	\section{Background and previous work}
	\paragraph{Adversarial examples:}
	Adversarial examples introduce minor perturbations of an input image, which makes a DNN classifier to fail \cite{Szegedy2013,Goodfellow2014}, also without access to the particular model \cite{Kurakin2016}, and by performing natural image operations \cite{Azulay2018,Engstrom2017,Zheng2016}. This points to the large sensitivity to the input of DNNs and, for image-to-image CNNs, it is manifested in inconsistent changes between frames when applied to a video sequence. Our goal is to train for robustness when it comes to the type of changes that can occur between frames in video sequences, so that video processed with CNNs can be expected to be well-behaved. This does not mean that the CNN will be robust to other types of changes, such as those created by certain adversarial example generation methods.
	
	\paragraph{Regularization:}
	While there exists a wide range of methods that classify as regularization \cite{Kukavcka2017}, we are particularly interested in those that are designed to address the issue of neural networks input sensitivity. Depending on the context and different definitions, the terms invariance, robustness, insensitivity, stability, and contraction have been used interchangeably in the literature for describing the objective of such regularization.
	
	The most straightforward method for increasing robustness and generalization is to employ data augmentation. However, augmentation alone cannot compensate for a CNN's sensitivity to transformations of the input \cite{Azulay2018,Engstrom2017} or degradation operations \cite{Zheng2016}. It would require too much training data to learn robustness for all transformations, and will most likely result in under-fitting. An explicit constraint needs to be enforced to learn a mapping that is smooth, so that small changes in input yield to small changes in output. This concept has been explored in a variety of formulations, \eg by means of weight decay \cite{Krogh1992}, weight smoothing \cite{Jean1994}, label smoothing \cite{Zhang2018}, or penalizing the norm of the output derivative with respect to the network weights \cite{Hochreiter1995}. Of particular interest to our problem are methods that regularize by penalizing the norm of the Jacobian with respect to the input \cite{Rifai2011,Zheng2016}. For example, Zheng \etal \cite{Zheng2016} apply noise perturbations to the input images, and construct a regularization term that contracts the prediction of clean and noisy samples, resulting in an increased robustness to image degradation.
	
	While the aforementioned works mostly deal with classification, we show that the same reasoning is true for image-to-image CNNs applied to video sequences --- we cannot simply train a CNN on separate video frames, or transformed images by means of augmentation, and expect 
	a robust behavior for temporal variations. Therefore, we formulate different regularization strategies particularly for training CNNs for video applications, and perform a study on which is most efficient for achieving temporal stability.
	
	\paragraph{Temporal consistency:}
	Methods for enforcing temporal consistency in image processing are mostly based on estimating dense motion, or optical flow, between frames \cite{Lang2012,Bonneel2015,Dong2015,Yao2017}. This is also the case for previous work in temporally consistent CNNs. For example, flow-based methods have been suggested for video style transfer \cite{Ruder2016,Gupta2017}, video-to-video operations by means of generative adversarial networks (GANs) \cite{Wei2018}, and for imposing temporal consistency as a post-processing operation \cite{Lai2018}. 
	
	Another direction for video inference using neural networks is to employ recurrent learning structures, such as the long short-term memory (LSTM) networks \cite{Hochreiter1997}. For image data, CNNs have been constructed for recurrence using the ConvLSTM \cite{Xingjian2015} and its variants \cite{Kalchbrenner2017}, which have been used \eg in video super-resolution \cite{Tao2017} and video concistency methods \cite{Lai2018}. However, mostly these structures have been explored in classification and understanding. There are also other recurrent or multi-frame based structures that have been used for image-to-image applications, \eg for video super-resolution \cite{Huang2015,Caballero2017}, de-blurring \cite{Su2017}, and different applications of GANs \cite{Wei2018}.
	
	The flow-based and recurrent methods all suffer from one or more of the following problems: 1) high complexity and application specific architectural modifications, 2) need for special-purpose training data such as video frames and motion information, 3) a significant increase in computational complexity for training and/or inference, 4) failure in situations where motion estimation is difficult, such as image regions with occlusion or lack of texture. The strategy we propose handles all these limitations. It is light-weight, can be applied to any image-to-image CNN without changes, and does not require video material or motion estimation. At the same time, it offers great improvements in temporal stability without impeding the reconstruction performance.
	
	
	\section{Temporal regularization}
	\label{sec:method}
	
	We consider supervised training of image-to-image CNNs, with the total loss formulated as:
	\begin{equation}
	\mathcal{L} = (1-\alpha)\mathcal{L}_{rec} + \alpha\mathcal{L}_{reg}.
	\label{eqn:loss}
	\end{equation}
	The first term is the main objective of the CNN, which promotes reconstruction of ground truth images $y$ from the input images $x$. Given an arbitrary CNN that has been trained with the loss $\mathcal{L}_{rec}$, adding the term $\mathcal{L}_{reg}$ is the only modification we make in order to adapt a CNN for video material. The scalar $\alpha$ is used to control the strength of the regularization objective.
	
	This section presents three different regularization strategies, $\mathcal{L}_{reg}$ in \eqnref{loss}, for improving temporal stability of CNNs. The first was introduced by Zheng \etal \cite{Zheng2016}, while the two others are novel definitions that are specifically designed to account for frame-to-frame changes in video. All three strategies rely on performing perturbations of the input image, and a key aspect is to model these as common transformations that occur in natural video sequences.
	
	
	\subsection{Stability regularization}\label{sec:stability}
	The most similar to our work is the stability training presented by Zheng \etal \cite{Zheng2016}. 
	Given an input image $x$, and a variant of it with a small perturbation $T(x) = x + \Delta{x}$, the regularization term is formulated to make the prediction of both images possibly similar. For an image-to-image mapping $f$, we can apply the term directly on the output image,
	\begin{equation}
	\mathcal{L}_{stability} = ||f(x)-f(T(x))||_2.
	\label{eqn:stability_loss}
	\end{equation}
	While different distance measures can be used, we only consider the $\ell_2$ norm for simplicity. The perturbation $\Delta{x}$ is described as per-pixel independent normally distributed noise, $\Delta{x} \sim \mathcal{N}(0,\Sigma)$, with $\Sigma = \sigma^2I$.
	
	
	
	\subsection{Transform invariance regularization}
	The typical measure of temporal incoherence \cite{Lang2012,Bonneel2015} is formulated using two consecutive frames $y_{t-1}$ and $y_t$,
	\begin{equation}
	E = ||y_t - W(y_{t-1})||_2,
	\end{equation}
	where $W$ describes a warping operation from frame $t-1$ to $t$ using the optical flow field between the two frames. If there are frame-to-frame changes that cannot be explained by the flow field motion, 
	these are registered as inconsistencies. 
	
	In order to use this measure for regularization, without requiring video data or optical flow information, we introduce within-frame warping with a geometric transformation, $W(x) = T(x)$ (the transformation is described in more detail in \secref{transf}). Then, $x$ and $T(x)$ mimic two consecutive frames, which are used to infer $f(x)$ and $f(T(x))$. If these are temporally consistent, performing the warping to register the two frames should yield the same result, either comparing $f(x)$ to $T^{-1}(f(T(x)))$ or comparing $T(f(x))$ to $f(T(x))$. This results in the regularization term:
	\begin{equation}
	\mathcal{L}_{trans{\text{-}}inv} = ||f(T(x)) - T(f(x))||_2\ .
	\label{eqn:warp_loss}
	\end{equation}
	Note that this loss is fundamentally different from the standard reconstruction loss for an augmented patch: 
	\begin{equation}
	\mathcal{L}_{augment} = ||f(T(x)) - T(y)||_2\ .
	\label{eqn:aug_loss}
	\end{equation}
	While $\mathcal{L}_{augment}$ promotes an accurate reconstruction with respect to an augmented (transformed) sample, $\mathcal{L}_{trans{\text{-}}inv}$ promotes the reconstruction that is consistent with a transformation, but not necessarily accurate. If there is an error in the reconstruction, $\mathcal{L}_{augment}$ will minimize that error in the transformed (augmented) patch, potentially at the cost of consistency, while $\mathcal{L}_{trans{\text{-}}inv}$ will ensure that any error is consistent between the original and the transformed patches. 
	
	\subsection{Sparse Jacobian regularization}
	Supervised learning typically relies on fitting a function to a number of training points without considering what is the function behavior in the neighborhood of those points. It would be arguably more desirable to provide to the training not only the function values, but also the information about partial derivatives in a form of a Jacobian of that function at a given point. However, for typical image-to-image CNNs, using a full Jacobian matrix would be impractical: if $32{\times}32$ patches are used, we need to train $f:\mathbb{R}^{1024}{\rightarrow}\mathbb{R}^{1024}$ and the Jacobian has over a million elements. However, we are going to demonstrate that even if we use a sparse estimate of the Jacobian and sample just a few random directions of our input space, we can much improve stability and accuracy of the predictions. 
	
	By providing sparse information on the Jacobian, we can also infuse domain expertise into our training. In the case of image-to-image mapping, we know that an input patch transformed by translation, rotation and scaling, should result in a transformed output patch. Each of those transformations maps to a vector change in the input and output space, for which we can numerically estimate partial derivatives. That is, we want the partial derivatives of the trained function $f$ to be possibly close to those of the ground truth output patches:
	\begin{equation}
	\frac{f(x+\Delta{x})-f(x)}{\Delta{x}} \approx \frac{y(x+\Delta{x})-y(x)}{\Delta{x}}\ ,
	\end{equation}
	where $\Delta{x}$ represents the effect of one of the transformations on the input space, $y(x)$ is the output patch from the training set corresponding to $x$, and $y(x+\Delta{x})$ is the transformed output patch. For the consistency of notation, we define $T(x)=x+\Delta{x}$ and $T(y)=y(x+\Delta{x})$, so that we can formulate a regularization term as:
	\begin{align}
	\mathcal{L}_{jacobian} &= || \left(f(T(x))-f(x)\right) - \left(T(y)-y\right)||_2 \\
	&=|| \left(f(T(x))-T(y)\right) - \left(f(x)-y\right)||_2\ .
	\label{eqn:jacobian_loss}
	\end{align}
	Although the term may look similar to $\mathcal{L}_{augment}$ from \eqnref{aug_loss}, $\mathcal{L}_{jacobian}$ promotes consistency rather than accuracy: the loss is minimized when the prediction error for the transformed patches is similar to the prediction error for the original patches. 
	
	

	\subsection{Transformation specification}\label{sec:transf}
	The perturbation function $T({\cdot})$ in all of the introduced regularization terms rely on a transformation of the input image.  
	For our purpose, this should capture the possible motion that can occur between frames in a video sequence. We make use of simple geometric transformations in order to accomplish this. These include translation, rotation, zooming, and shearing, which all can be described in a $2\times3$ transformation matrix that transforms the indices of the image $x$ \ADD{(see supplementary material for an exact formulation)}. The matrix is randomly specified for each image, with transformation parameters drawn from uniform distributions in a selected range of values as specified in \tabref{transf_values}.

	\begin{table}[t!]
		\centering
		\setlength{\tabcolsep}{15pt}
		\def\arraystretch{0.95}
		\small
		\caption{Ranges of transformation parameters.}
		\vspace{0.2cm}
		\begin{tabular}{l|l|l}
			\textbf{Parameter} & \textbf{Min} & \textbf{Max}\\
			\hline
			Translation & -2 px & 2 px \Tstrut\\
			Rotation & -1$^\circ$ & 1$^\circ$\\
			Zoom & 0.97$\times$ & 1.03$\times$\\
			Shearing & -1$^\circ$ & 1$^\circ$\\
		\end{tabular}
		\label{tab:transf_values}\\
		\vspace{-0.2cm}
	\end{table}
	
	\ADD{Although motion can occur on a more local level in real videos, we argue that the transformations can make for good regularization. It is important to note that we do not train the network to predict the transformation or transformed patch, which would arguably require training on local transformations. Instead, we train to produce an image in which pixels do not shift and which is consistent with the input in the presence of any transformation, both local or global. We argue that the type of motion (global/local) is in this case less relevant as long as the regularization term pushes the trained model towards predicting consistent results.}
	
	
	
	\subsection{Implementation}
	While it is possible to train for a loss function with one of the regularization terms from scratch, we instead start with a pre-trained network and include the regularization in a second training stage for fine-tuning. We found that fine-tuning makes training convergence more stable while providing the same gain in temporal consistency as training from scratch. Another very important advantage is that fine-tuning can be applied to already optimized large-scale CNNs, which take long time to train.
	
	For each regularization method, we follow the exact same loss evaluation scheme. The perturbed sample's coordinates are transformed as described in \secref{transf}, with randomly selected transformation parameters. Both the original and the transformed sample, $x$ and $T(x)$, respectively, are taken through the CNN by the means of a weight-sharing (siamese) architecture. This gives us $f(x)$ and $f(T(x))$, which can be used with the three different regularization definitions, \eqnref{stability_loss}, \ref{eqn:warp_loss}, and $\ref{eqn:jacobian_loss}$, by complementing with the transformations $T(f(x))$ and $T(y)$.
	
	
	
	\section{Experiments}\label{sec:experiment}
	
	We evaluate the novel temporal CNN stabilization/regularization techniques using two different applications: colorization of grayscale images and HDR reconstruction from single-exposure images. These tasks were selected as they are different in nature, and rely on different CNN architectures. While colorization attempts to infer colors over the complete image, the HDR reconstruction tries to recover local pixel information that have been lost due to sensor saturation.
	The colorization CNN uses the same design as described by Iizuka \etal \cite{Iizuka2016}, but without the global features network and with fewer weights. It implements an autoencoder architecture, with strided convolution for down-sampling, and nearest neighbor resizing followed by convolution for up-sampling. The HDR reconstruction CNN uses the same design as described by Eilertsen \etal \cite{Eilertsen2017}, but with fewer weights. This is also an autoencoder architecture, but implemented using max-pooling and transposed convolution, and it has skip-connections between encoder and decoder networks. More details on the CNNs and training setups are listed in \tabref{cnns}.
	
	\begin{table}[t!]
		\centering
		\setlength{\tabcolsep}{4pt}
		\def\arraystretch{0.95}
		\small
		\caption{CNN training setups used in the evaluation experiments.}
		\vspace{0.2cm}
		\begin{tabular}{l|l|l}
			& \textbf{Colorization} & \textbf{HDR reconstruction}\\
			\hline
			Architecture & Autoencoder \cite{Iizuka2016} & Autoencoder \cite{Eilertsen2017} \Tstrut\\
			Down-sampling & Strided conv. & Max-pooling\\
			Up-sampling & Resize + conv. & Transposed conv.\\
			Skip-connections & No & Yes \\
			Weights & 1,568,698 & 1,289,653\\
			Training data & CelebA \cite{Liu2015} & Procedural images\\
			Resolution & $128\times128$ & $128\times128$\\
			Training size & 20,000 & 10,000\\
			Epochs & 50 & 50\\
			Training time & $\approx$35m & $\approx$20m \\
		\end{tabular}
		\label{tab:cnns}
		\vspace{-0.2cm}
	\end{table}
	
	In order to be able to explore a broad range of hyper-parameters, we use datasets that are restricted to specific problems. For colorization, we only learn the task for close-up face shots. For the HDR reconstruction, we restrict the task to a simple procedural HDR animation.
	
\begin{figure}[t!]
		\centering
		\includegraphics[width=\linewidth]{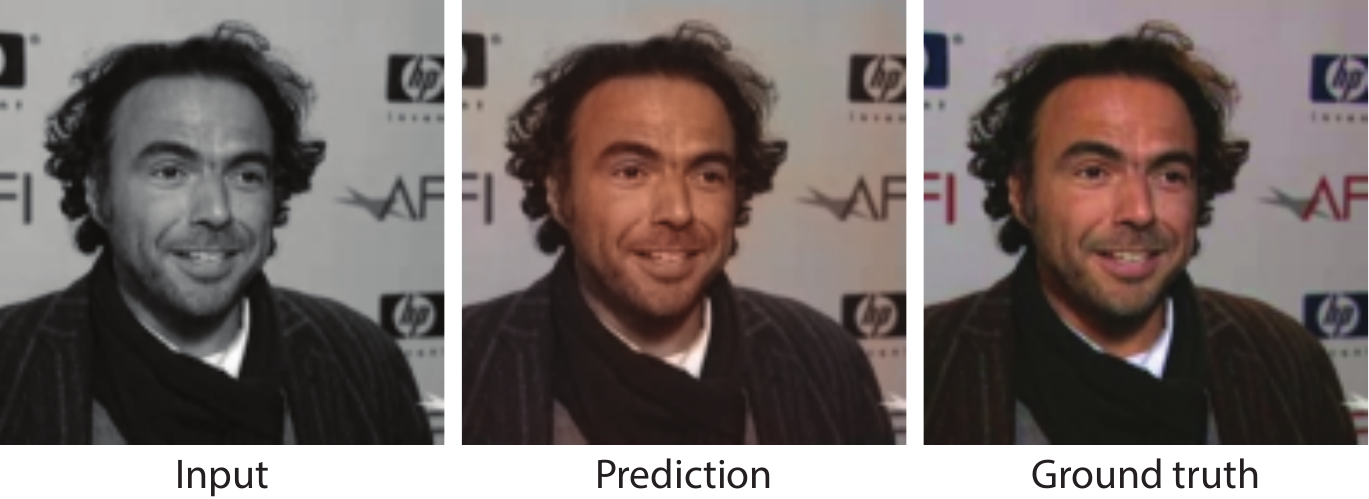}
		\caption{\label{fig:col_samp} Colorization test sample, from YouTube Faces \cite{Wolf2011}.}
	\end{figure}

	Training data for the colorization task is 20,000 images from the CelebA dataset \cite{Liu2015}. For testing, we use 72 video sequences from the YouTube Faces dataset \cite{Wolf2011}. These have been selected to show close-up faces in order to be more similar to the training data, and are cut to be between $50-200$ frames long. \figref{col_samp} shows an example of a test frame.
	
\begin{figure}[t!]
		\centering
		\includegraphics[width=\linewidth]{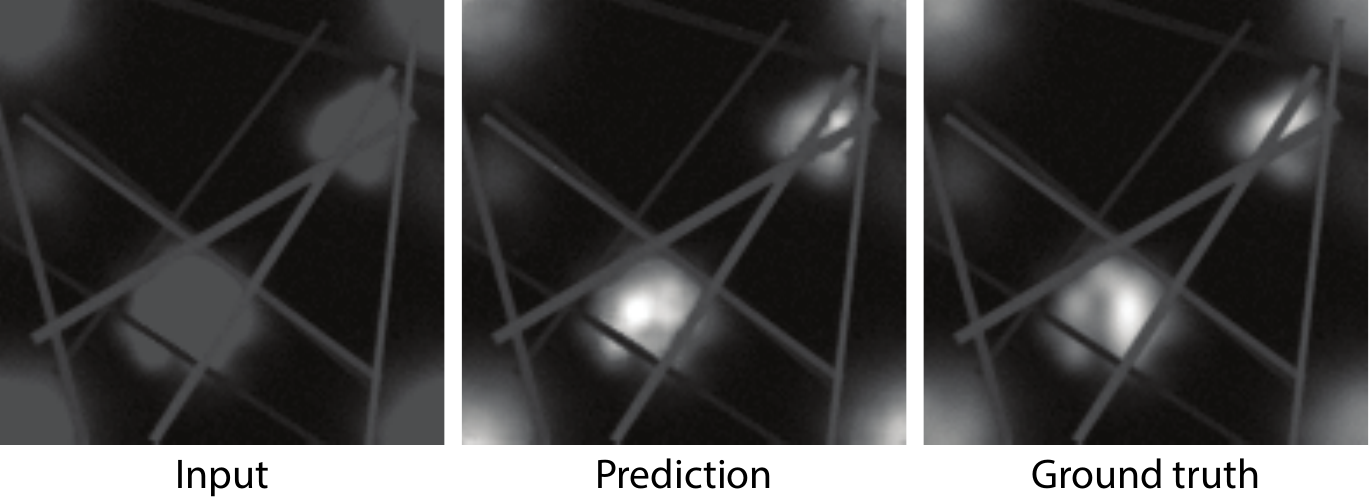}
		\caption{\label{fig:hdr_samp} HDR reconstruction test sample, from procedural HDR video dataset. The image is displayed at a reduced exposure in order to show differences in saturated regions.}
	\end{figure}

	Training data for the HDR reconstruction task is 10,000 frames that have been generated in a completely procedural manner. These contain a random selection of image features with different amount of saturated pixels. The features move in random patterns and are sometimes occluded by randomly placed beams. For the training data we only use static images, with no movement, and for the test data we include motion to evaluate the temporal behavior. The test set consists of 50 sequences, 200 frames each. \figref{hdr_samp} shows an example of a test video frame.
	
	\subsection{Performance measures}
\begin{figure}[t!]
		\centering
		\includegraphics[width=0.9\linewidth]{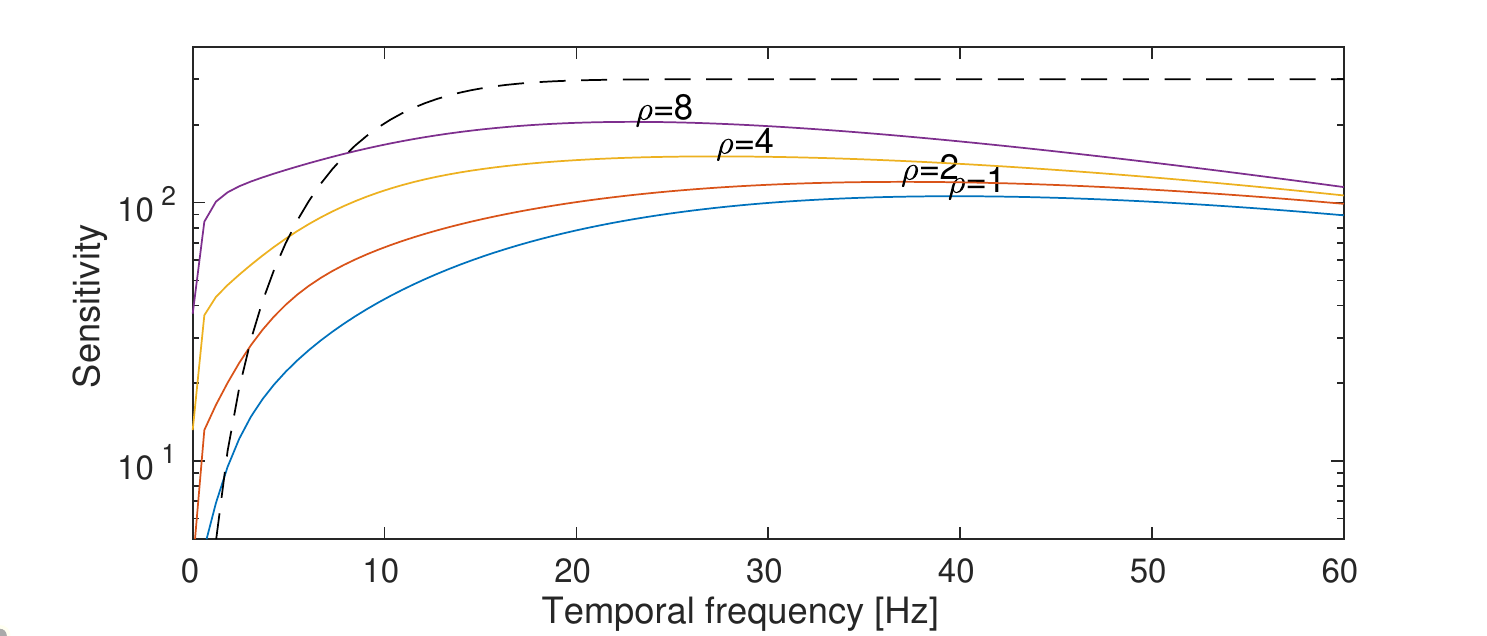}
		\caption{\label{fig:spatio-temp-csf} Colored lines: Spatio-temporal contrast sensitivity function for different spatial frequencies $\rho$ in cycles per visual degree, based on the model from \cite{Laird2006}. Dashed-black line: the high-pass filter used for our smoothness measure.}
	\end{figure}

	The goal of the proposed regularization strategies is to achieve temporally stable results while maintaining the reconstruction performance. In order to evaluate whether both goals are achieved, we measure reconstruction performance by means of PSNR and introduce a new measure of smoothness over time. Our measure computes the ratio of high temporal frequencies between the reference and reconstructed video sequences. We first extract the energy of the high temporal frequency component from both sequences, 
	\begin{equation}
	D(f(x))_{i,j,t} = |f(x)_{i,j,t} - (G_\sigma * f(x))_{i,j,t}|^2\ ,
	\end{equation}
	where the convolution with the Gaussian filter $G_\sigma$ is performed in the temporal dimension $t$. The parameter $\sigma$ is selected to eliminate the low frequency components that the eye is insensitive to, but which carry high energy. Figure~\ref{fig:spatio-temp-csf} shows the spatio-temporal contrast sensitivity function of the visual system and the high-pass filter we use with $\sigma=0.15$\;seconds. The smoothness is computed as the ratio of the sum of the ground truth and the reconstruction video energies,
	\begin{equation}
	S = \sqrt{\frac{\sum_{i,j,t} D(y)_{i,j,t}}{\sum_{i,j,t} D(f(x)_{i,j,t})}}.
	\end{equation}
	If $S<1$, the reconstructed video is less smooth than the ground truth video and the opposite can be said for $S>1$.
	
	\subsection{Experimental setup}
	We fine-tune the CNNs in \tabref{cnns} for the two applications, and run a large number of trainings in order to sample the performance at different settings. For the total loss in \eqnref{loss}, we compare the three different regularization formulations: stability \eqref{eqn:stability_loss}, sparse Jacobian \eqref{eqn:jacobian_loss}, and transform invariance \eqref{eqn:warp_loss}. These are evaluated using the transformation described in \secref{transf}. For the stability regularization we also include a setting with noise perturbations, $T(x) = x + \Delta{x}$, with $\Delta{x} \sim \mathcal{N}(0,\sigma^2I)$, in order to compare to previous work. We choose different $\sigma$ for each image, drawn from a uniform distribution, $\sigma \sim \mathcal{U}(0.01,0.04)$. Finally, we also include trainings that use traditional augmentation by means of the transformation. For each of the aforementioned setups, we then run 10 individual trainings in order to estimate the mean and standard deviation of each datapoint.
	
	We also experimented with incorporating the reconstruction loss of the transformed sample, \eqnref{aug_loss}, but mostly this degraded the performance, possibly due to under-fitting.
	
	\subsection{Results}
\begin{figure}[t!]
		\newcommand\figw{0.75\linewidth}
		\centering
		\includegraphics[width=\figw]{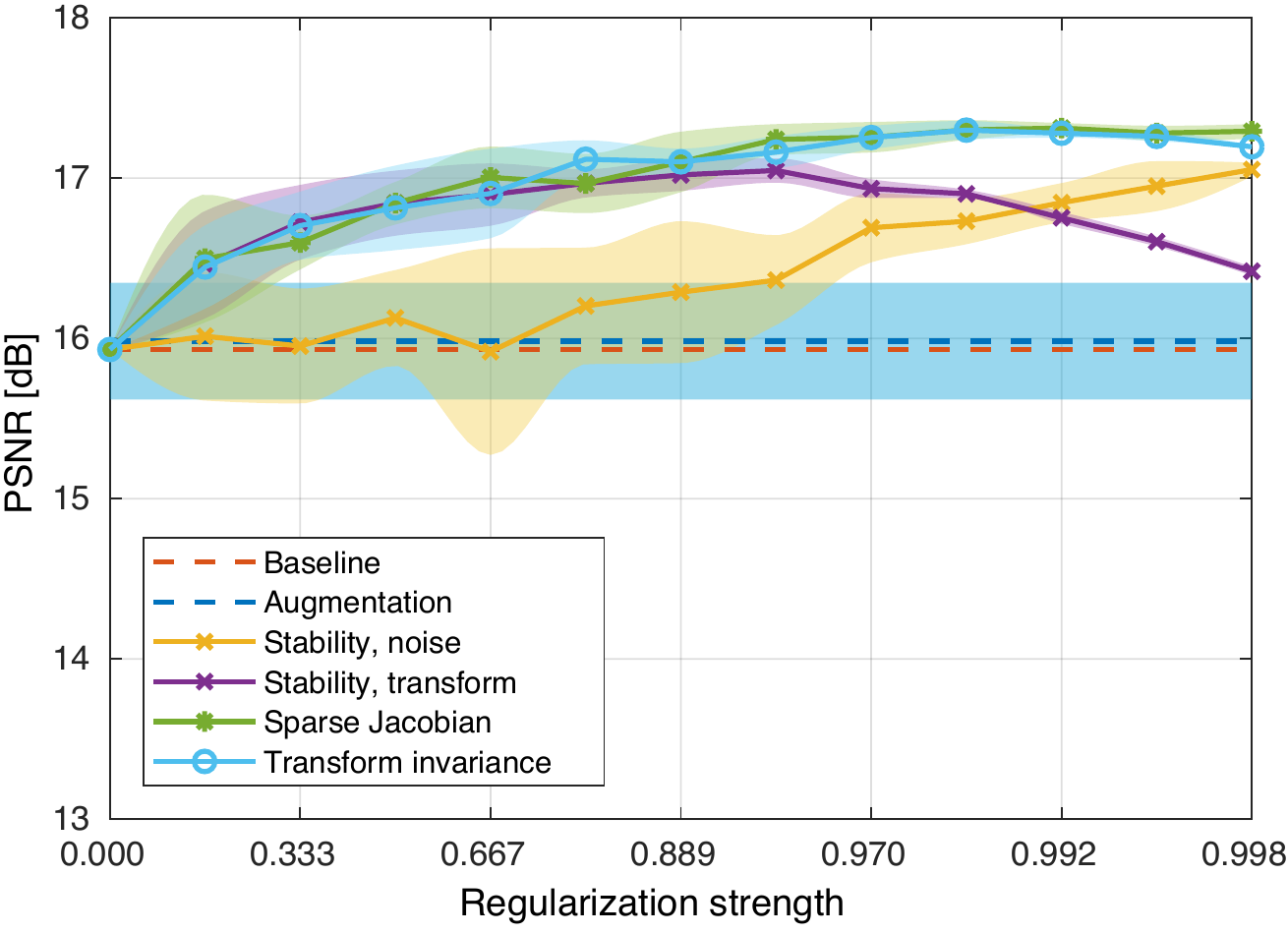}
		\vspace{3pt}\\
		\includegraphics[width=\figw]{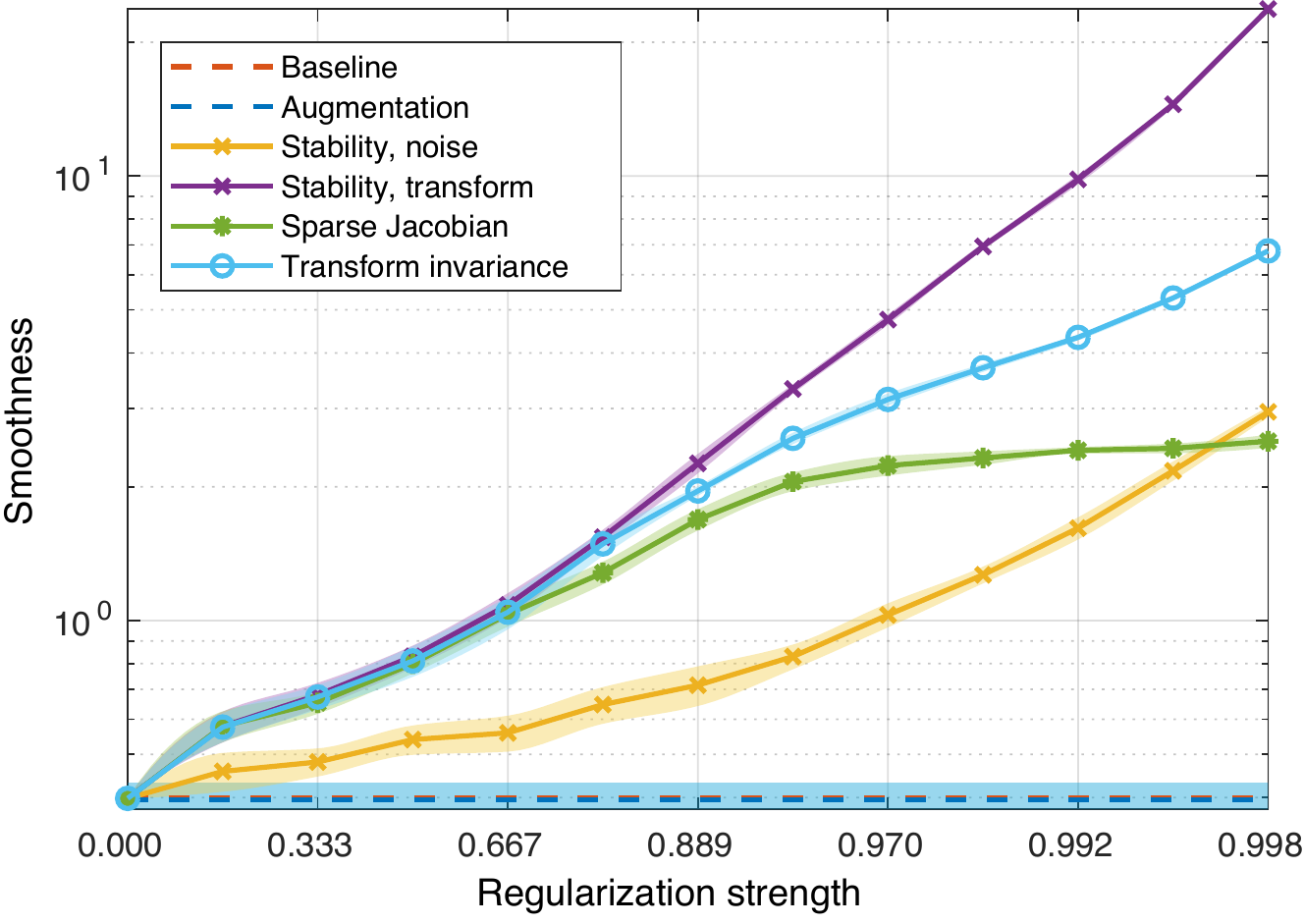}
		\caption{\label{fig:exp_col} Colorization performance, evaluated using PSNR (top) and smoothness (bottom). The datapoints are estimated as the mean over 10 individual trainings, and the shaded regions illustrate the standard deviation. The Baseline condition is hidden under the dashed-blue Augmentation line in the bottom plot.}
	\end{figure}

\begin{figure}[t!]
		\newcommand\figw{0.75\linewidth}
		\centering
		\includegraphics[width=\figw]{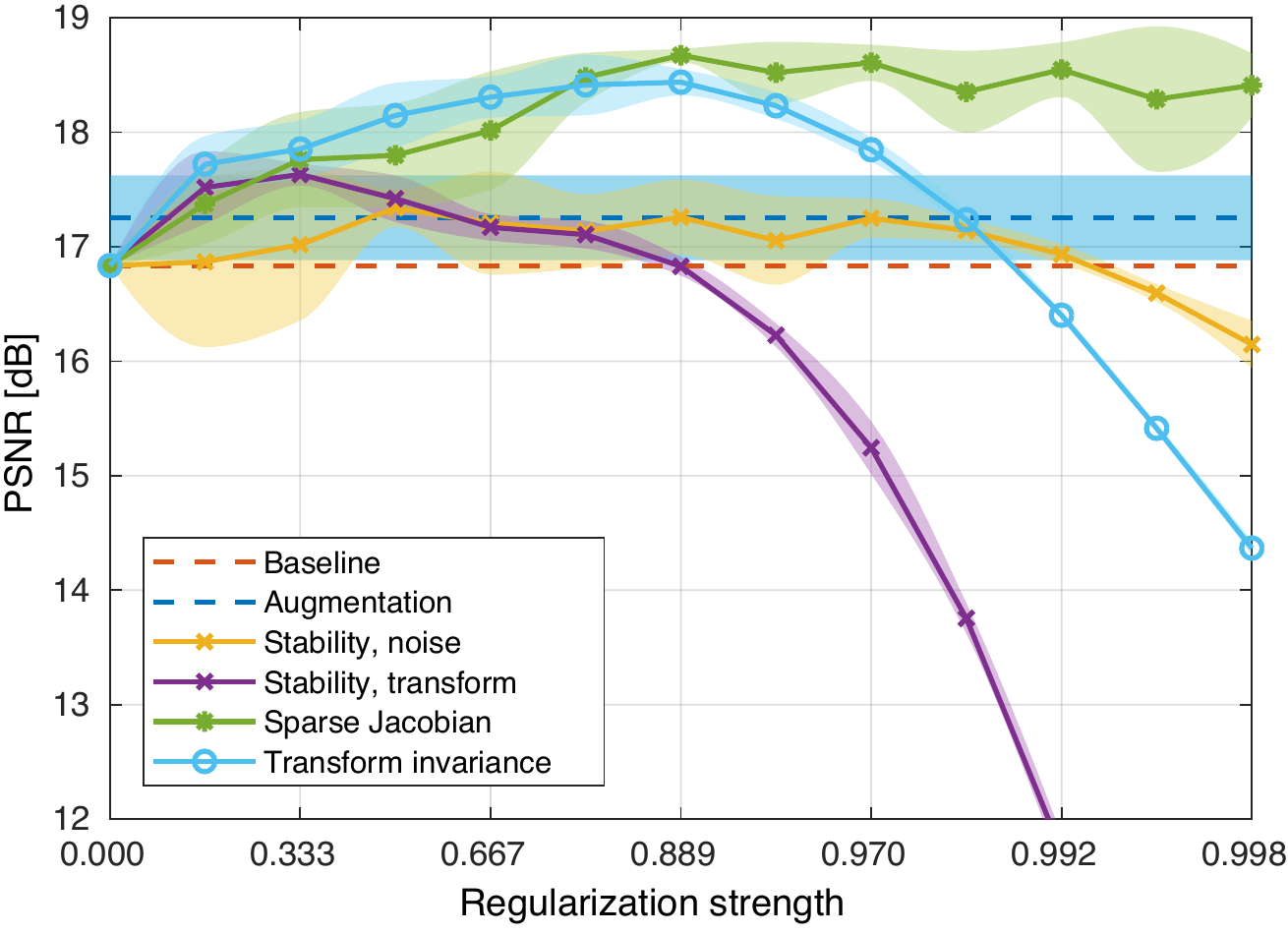}
		\vspace{3pt}\\
		\includegraphics[width=\figw]{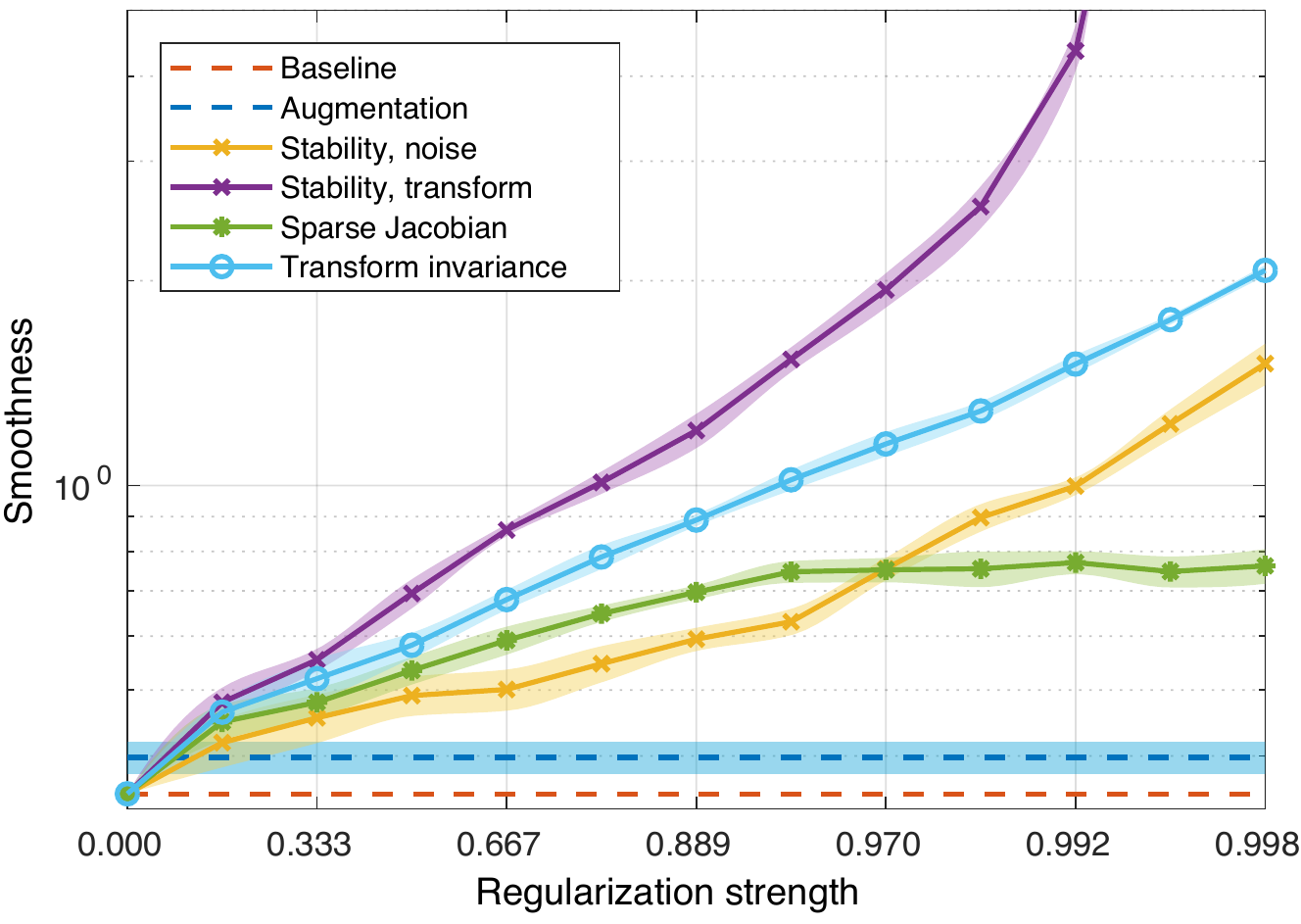}
		\caption{\label{fig:exp_hdr} HDR reconstruction performance, evaluated using PSNR (top) and smoothness (bottom). The notation is the same as in \figref{exp_col}.
		}
	\end{figure}

	The results of the experiments can be found in \figref{exp_col} for colorization and in \figref{exp_hdr} for HDR reconstruction. The baseline condition uses the pre-trained model before fine-tuning and without regularization. The PSNR and smoothness measures have been calculated on the $a$ and $b$ channels of the CIE~Lab color space for the colorization application and only in saturated pixels for the HDR reconstruction application. Such modified measures can better capture small differences. 
	
	In both experiments we can observe significant improvements in both PSNR and smoothness for all regularization strategies. However, the \emph{stability} that relies on noise performs visibly worse in both experiments than the same regularization but based on transformations. \emph{Transform invariance} and \emph{sparse Jacobian} regularizations result in higher PSNR and visually better reconstruction than the \emph{stability} regularization (refer to the video material). Although the stability formulation can generate smoother video for HDR reconstruction, this is at the cost of very high reconstruction error, and for $\alpha > 0.99$ it most often learns the identity mapping, $f = x$. The performance of the two novel formulations are comparable. The sparse Jacobian results in a slightly higher PSNR for HDR reconstruction and transform invariance results in higher smoothness. The sparse Jacobian also seems to be more robust to the choice of the regularization strength. The traditional augmentation using the transformations (the blue-dashed line) can improve smoothness and PSNR but the improvement is much smaller than the other regularization strategies.

	In summary, the experiments give us a good indication of the large improvements in temporal stability for widely different applications that can be achieved from explicitly regularizing for this objective. 
	However, differentiating between the two proposed formulations is more difficult, and could potentially be application dependent.
	Finally, we have large improvements in PSNR for our scenarios with limited training data, indicating that the proposed regularization strategies can improve generalization performance. 
	

	\section{Example applications}
	In  this section we demonstrate that the proposed regularization terms improve the results not only for the limited scenarios in \secref{experiment}, but also for large-scale problems trained on large amounts of data. 
	
	
	\subsection{Colorization}
	\begin{figure*}[t!]
		\centering
		\includegraphics[width=0.9\linewidth]{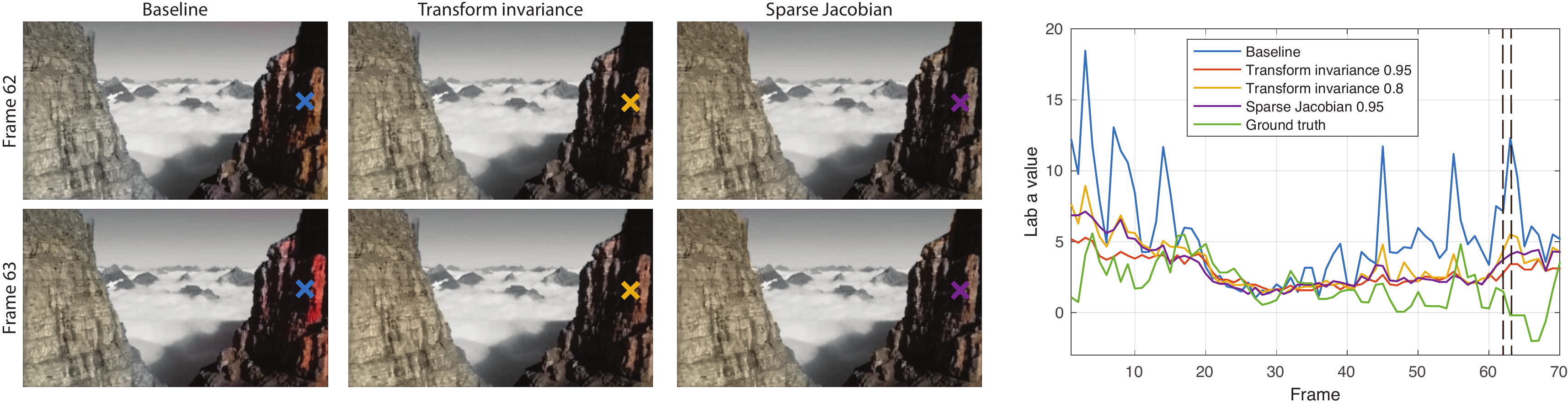}\\
		\vspace{0.1cm}
		\includegraphics[width=0.9\linewidth]{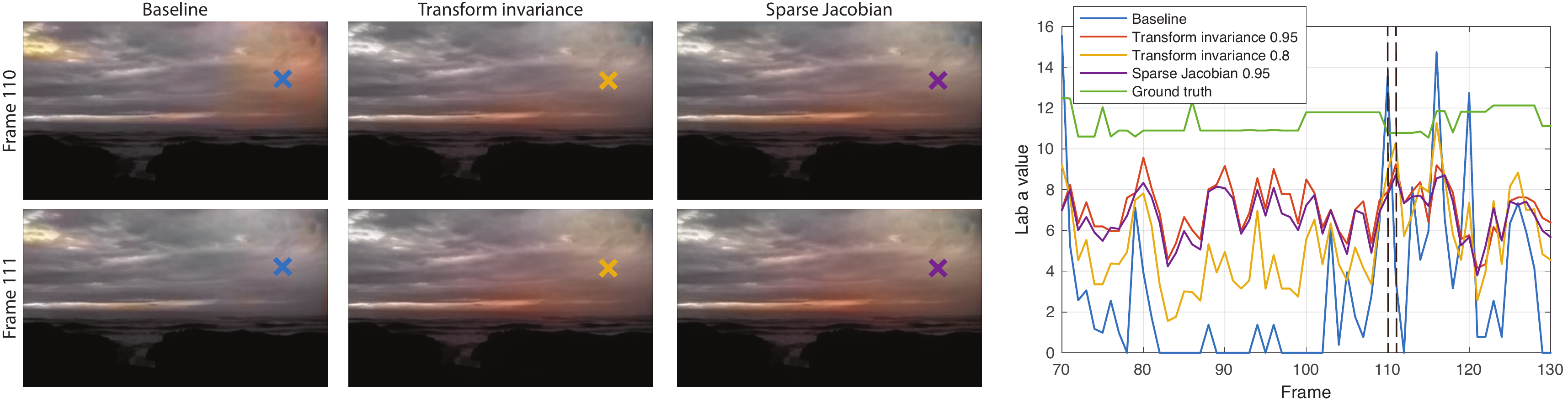}
		\caption{\label{fig:col3} Two video colorization examples from the YouTube-8M dataset \cite{Abu2016}. On the left there are two consecutive frames displayed for each sequence, comparing baseline to the two video regularization techniques. The plots on the right show the pixel values of the locations marked in the frames, over a larger range of frames. The values are taken from the $a$ channel of the Lab color space. The vertical dashed lines indicate where the displayed frames are. The transform invariance regularization has been performed at two strengths, $\alpha$.}
	\end{figure*}
	
	
	For this application, we start from the architecture used by Iizuka \etal \cite{Iizuka2016}. However, we skip the global features network and replace the encoder part of the CNN with the convolutional layers from VGG16 \cite{Simonyan2014}. In this way, we can initialize the encoder using pre-trained weights for classification. This setup resulted in a significant improvement in the performance as compared to using the original encoder design. In total, the network is specified from $\sim$19M weights. We train it on the Places dataset \cite{Zhou2014}, and use weights pre-trained for classification on the same dataset. We remove from training around 5\% of the images that showed the least color saturation. The CNN was then trained for $\sim$15 epochs on the remaining $\sim$2.1M images, at a resolution of $224\times224$ pixels.
	
	We fine-tune the colorization CNN using two proposed regularization strategies. The effect of the fine-tuning is measured in terms of PSNR and the smoothness measure, see \tabref{ft_col}. The table also includes a fine-tuning without regularization for comparison, and processing the baseline output using the method by Lai \etal \cite{Lai2018}. Overall, the regularizations offer slight improvements in PSNR (around 0.3$-$0.5dB) while increasing smoothness substantially. This also goes for comparison to the flow-based post-processing network by Lai \etal. The transform invariance formulation with $\alpha = 0.95$ gives the best smoothness, and with a PSNR close to the other regularization settings.
	
	\begin{table}[!t]
		\centering
		\setlength{\tabcolsep}{6pt}
		\def\arraystretch{0.95}
		\small
		\caption{Performance after fine-tuning of the colorization CNN. The measures have been evaluated and averaged over the $a$ and $b$ channels in the Lab color encoding. Test data are 23 sequences from the YouTube-8M dataset \cite{Abu2016}.}
		\vspace{0.2cm}
		\begin{tabular}{l|l|l}
			\textbf{Training strategy} & \textbf{PSNR} & \textbf{Smoothness} \\
			\hline
			Baseline & 18.5805 & 0.7243 \Tstrut\\
			Fine-tuning (no regularization) & 18.4315 & 0.6348\\
			Transform invariance, $\alpha=0.95$ & 18.8880 & \textbf{2.8934}\\
			Transform invariance, $\alpha=0.8$ & \textbf{18.9437} & 1.9074\\
			Sparse Jacobian, $\alpha=0.95$ & 18.8852 & 2.5079\\
			Blind video consistency \cite{Lai2018} & 18.6086 & 1.0287 \\
		\end{tabular}
		\label{tab:ft_col}
		\vspace{-0.2cm}
	\end{table}
	
	Examples of the impact of the regularization techniques are demonstrated in \figref{col3}. The baseline CNN can exhibit large frame to frame differences, which is much less likely after performing the regularized training. Also, there is an overall increase in the reconstruction performance --- whereas the baseline has a tendency to fail in many of the frames, this is less likely to happen when accounting for the differences between frames in the loss evaluation. For example, in the bottom example of \figref{col3} the pixel values plotted for the baseline CNN are in many cases close to 0, and occasionally spike to high values. This problem is alleviated by the regularization, resulting in both overall better reconstruction and smoother changes between frames.
	
	
	\subsection{HDR reconstruction}
	\begin{figure*}[t!]
		\centering
		\includegraphics[width=0.9\linewidth]{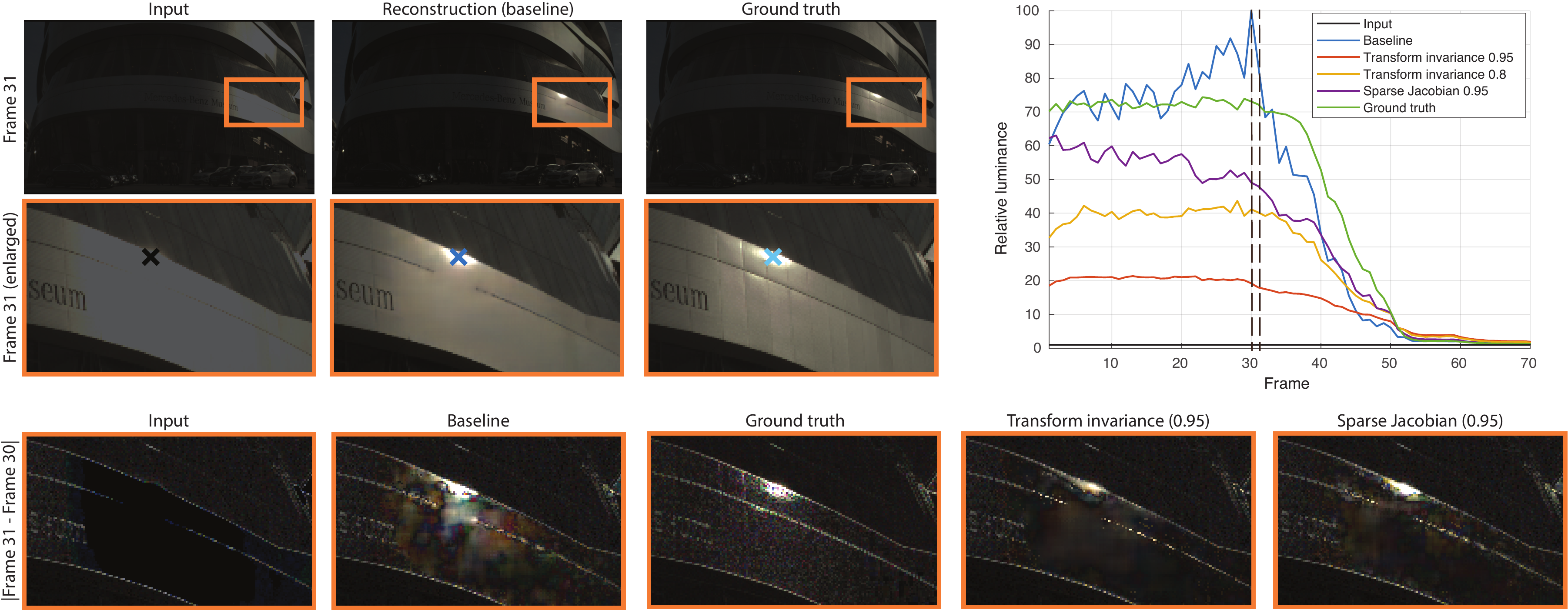}
		\vspace{0.1cm}
		\caption{\label{fig:hdr2} HDR video reconstruction example from the HdM-HDR dataset \cite{Froehlich2014}. On the top left we have an example of the reconstruction compared to input and ground truth, displayed at a reduced exposure ($-$3 stops) in order to demonstrate the differences in saturated pixels. The bottom row displays the absolute difference between two consecutive frames, for an enlarged region of the image and for different training strategies. The plot on the right shows the HDR luminance values of the pixels marked in the frames, over a larger range of frames. The vertical dashed lines indicate the frames used for the difference evaluations.}
	\end{figure*}
	
	In this application we employ the CNN that was used by Eilertsen \etal \cite{Eilertsen2017} and initialize it with the trained weights provided by the authors. The CNN contains in total $\sim$29M weights. We perform fine-tuning on a gathered set of $\sim$2.7K HDR images from different online resources, which are used to create a dataset of $\sim$125K $320\times320$ pixel training images by means of random cropping and augmentation.
	
	The fine-tuning result is measured by PSNR and smoothness in \tabref{ft_hdr}, demonstrating a significant increase in smoothness at the cost of a small decrease in PSNR. Compared to the colorization application, regularization of the HDR reconstruction should be selected at a slightly lower $\alpha$ in order to not degrade reconstruction performance. The transform invariance formulation at $\alpha = 0.8$ only reduces the reconstruction performance by $\sim$0.1dB while providing better smoothness than the sparse Jacobian formulation. This setting also shows better performance as compared to the blind video consistency method by Lai \etal \cite{Lai2018}, both in terms of PSNR and smoothness.
	
	\begin{table}[t!]
		\centering
		\setlength{\tabcolsep}{6pt}
		\def\arraystretch{0.95}
		\small
		\caption{Performance after fine-tuning of the HDR reconstruction CNN. The measures have been evaluated and averaged over the saturated pixels only. Test data are 10 HDR video sequences from two different sources \cite{Froehlich2014,Banitalebi2014}. The blind video consistency has been performed on gamma corrected HDR images.}
		\vspace{0.2cm}
		\begin{tabular}{l|l|l}
			\textbf{Training strategy} & \textbf{PSNR} & \textbf{Smoothness} \\
			\hline
			Baseline & 25.5131 & 5.9951 \Tstrut\\
			Fine-tuning (no regularization) & \textbf{25.9865} & 5.8538\\
			Transform invariance, $\alpha=0.95$ & 24.1678 & \textbf{10.6435}\\
			Transform invariance, $\alpha=0.8$ & 25.4374 & 8.0798\\
			Sparse Jacobian, $\alpha=0.95$ & 24.7287 & 7.3048\\
			Blind video consistency \cite{Lai2018} & 25.3702 & 7.2035 \\
		\end{tabular}
		\label{tab:ft_hdr}
		\vspace{-0.3cm}
	\end{table}
	
	\figref{hdr2} shows an example of the difference in performance for one HDR video sequence. In contrast to the colorization application it is difficult to clearly see the differences between consecutive frames in a side-by-side comparison. However, in the video material the differences in the temporal robustness around saturated image regions are evident. This can be seen in the pixel plots in \figref{hdr2}, where the regularized results are more stable over time for the selected saturated pixel. The figure also shows the absolute difference between two frames for an enlarged image region, highlighting the improvements achieved from regularization when comparing to the ground truth difference.
	
	
	
	
	\section{Limitations and future work}
	Striking the right balance between reconstruction performance and smoothness is still an open problem. A small regularization strength leaves video with temporal artifacts, whereas a too large strength may risk degrading the reconstruction performance. 
	\ADD{Also, the tendency to impair reconstruction performance with strong regularization could be in some respect analogous to the reduced sharpness when L2 norm is used as the loss function in reconstruction problems (denoising, deconvolution, etc.). We do not address this problem in our current work, but believe that this can be alleviated by exploring other regularization loss functions, such as L1, perceptual loss (for color), or by means of a GAN architecture.}
	The method could also benefit from combining the reconstruction error and smoothness, for a better measure of perceived quality. Moreover, although the transform invariance formulation in some situations can give a better trade-off between PSNR and smoothness, the sparse Jacobian formulation tends to be more robust to large regularization strengths, see \eg \figref{exp_hdr}.
	
	

	Our approach optimizes towards short-term temporal stability without a guarantee for the long-term temporal consistency. For example, even if colors are consistent in consecutive frames for the colorization application, they may change inconsistently over a longer sequence of frames. An interesting area for future work is therefore to investigate how long-term temporal coherence can be enforced upon the solution. 
	Finally, it would also be interesting to explore regularization of more complicated loss functions, such as those based on GANs \cite{Goodfellow2014b}, \eg the pix2pix \cite{Isola2017} CNN or cycle-GANs \cite{Zhu2017}.
	
	\section{Conclusion}
	This paper explored how regularization using models of the problem dynamics can be used to improve the temporal stability of pixel-to-pixel CNNs in video reconstruction tasks. We proposed two formulations for temporal regularization, which can be used when training a network from scratch, or for fine-tuning pre-trained networks. The strategy is light-weight, it can be used without architectural modifications of the CNN, and it does not require video or motion information for training. It avoids the costly and often inaccurate estimation of optical flow, inherent to previous stabilization methods. Our experiments showed that the proposed approach leads to substantial improvements in temporal stability while maintaining the reconstruction performance. 
	Moreover, for some situations, and especially when training data is limited, the regularization can also improve the reconstruction performance of the CNN, and to a much larger extent than what is possible with traditional augmentation techniques.
	
	\paragraph{Acknowledgments}
	{\small This project was supported by the Wallenberg Autonomous Systems and Software Program (WASP), the strategic research environment ELLIIT, and has received funding from the European Research Council (ERC) under the European Union's Horizon 2020 research and innovation programme (grant agreement n$^\circ$ 725253--EyeCode).}
	
	
	
	{\small
		\bibliographystyle{ieee}
		\bibliography{egbib}
	}

	\clearpage
	\newpage

\renewcommand\thesection{\Alph{section}}


\section*{\Large Supplementary material}
\vspace{0.2cm}
\section*{S1: Transformations}\label{sec:transf_app}
The image perturbations $T({\cdot})$ are performed by means of a linear transformation of the pixel indices $i$ and $j$,
\begin{equation}
\begin{bmatrix}
i' \\ j'
\end{bmatrix}
=
\begin{bmatrix}
T_{1,1} & T_{1,2} & T_{1,3}\\
T_{2,1} & T_{2,2} & T_{2,3}
\end{bmatrix}
\begin{bmatrix}
i \\ j \\ 1
\end{bmatrix},
\label{eqn:transf}
\end{equation}
where $(i',j')$ are the transformed indices, such that $T(x)_{i,j} = x_{i',j'}$. The transformation matrix elements are defined as follows:
\begin{equation}
\begin{split}
T_{1,1} &= \frac{z\cos a}{\cos(h_x)},\\
T_{1,2} &= \frac{z \sin(a)}{\cos(h_x)},\\
T_{1,3} &= \frac{s_x\cos(h_x) - s_xz\cos a}{2\cos(h_x)} \\
&+ \frac{2t_xz\cos a - s_yz\sin a + 2t_yz\sin a}{2\cos(h_x)}, \\
T_{2,1} &= \frac{z\sin b}{\cos(h_y)},\\
T_{2,2} &= \frac{z\cos b}{\cos(h_y)}, \\
T_{2,3} &= \frac{s_y\cos(h_y) - s_yz\cos b}{2\cos(h_y)} \\
&+ \frac{2t_yz\cos b - s_xz\sin b + 2t_xz\sin b}{2\cos(h_y)}.
\end{split}
\end{equation}
Here, we have $a=h_x-r$ and $b=h_y+r$, and $(s_x,s_y)$ is the image size. The formulation assumes that the image origin is in the corner of the image, thus incorporating a translation of the image center to the origin before performing the image transformations and translating back afterwards. $(t_x,t_y)$, $r$, $z$, and $(h_x,h_y)$ are translation offset, rotation angle, zoom factor, and shearing angles, respectively. All the transformation parameters are drawn from uniform distributions, in a selected range of values as specified in \tabref{transf_values}.


\section*{S2: Implementation}
The transformations and the loss formulations in \secref{method} can be implemented with little modification of an existing CNN training script. An example implementation is provided in Listing \ref{lst:tf}, using Tensorflow. It evaluates the CNN on the input image $x$ and the transformed image $T(x)$ by means of a weight-sharing network.

\begin{lstlisting}[language=Python, caption={Tensorflow example for formulating regularized loss.}, label={lst:tf}, basicstyle=\tiny,float=tp,floatplacement=tbp]
import numpy as np
import tensorflow as tf

# Other initialization stuff
...

# The ground truth (bs is batch size)
y = tf.placeholder(tf.float32, [bs, sx, sy])

# Random transformations
ang = np.deg2rad(1.0)
tx = tf.random_uniform(shape=[bs,1], minval=-2.0, maxval=2.0, dtype=tf.float32)
ty = tf.random_uniform(shape=[bs,1], minval=-2.0, maxval=2.0, dtype=tf.float32)
r = tf.random_uniform(shape=[bs,1], minval=-ang, maxval=ang, dtype=tf.float32)
z = tf.random_uniform(shape=[bs,1], minval=0.97, maxval=1.03, dtype=tf.float32)
hx = tf.random_uniform(shape=[bs,1], minval=-ang, maxval=ang, dtype=tf.float32)
hy = tf.random_uniform(shape=[bs,1], minval=-ang, maxval=ang, dtype=tf.float32)

# Transformation matrix
a = hx - r
b = hy + r
T1 = tf.divide(z*tf.cos(a), tf.cos(hx))
T2 = tf.divide(z*tf.sin(a), tf.cos(hx))
T3 = tf.divide(sx*tf.cos(hx)-sx*z*tf.cos(a)+2*tx*z*tf.cos(a)-sy*z*tf.sin(a)+2*ty*z*tf.sin(a), 2*tf.cos(hx))
T4 = tf.divide(z*tf.sin(b), tf.cos(hy))
T5 = tf.divide(z*tf.cos(b), tf.cos(hy))
T6 = tf.divide(sy*tf.cos(hy)-sy*z*tf.cos(b)+2*ty*z*tf.cos(b)-sx*z*tf.sin(b)+2*tx*z*tf.sin(b), 2*tf.cos(hy))
T7 = tf.zeros([bs,2], 'float32')
T = tf.concat([T1, T2, T3, T4, T5, T6, T7], 1)

# Perform transformation
Ty = tf.contrib.image.transform(y, T, interpolation='BILINEAR')

# Prepare input x from ground truth y
x = prepare_data(y)
Tx = prepare_data(Ty)

# Model
with tf.variable_scope("siamese") as scope:
  fx = cnn_model(x)
  
  # Weight-sharing
  scope.reuse_variables()
  fTx = cnn_model(Tx)

# Transformation on prediction
Tfx = tf.contrib.image.transform(fx, T, interpolation='BILINEAR')

# Reconstruction loss
loss = (1.0-alpha)*tf.reduce_mean(tf.square(fx-y))

# Regularization loss
if stability:
  loss += alpha*tf.reduce_mean(tf.square(fx-fTx))
elif transform_invariance:
  loss += alpha*tf.reduce_mean(tf.square(fTx-Tfx))
elif sparse_jacobian:
  loss += alpha*tf.reduce_mean(tf.square((fTx-fx)-(Ty-y)))

# Train model using the regularized loss
...
\end{lstlisting}

\section*{S3: Training time}
The regularized losses take approximately 2 times longer to evaluate as compared to training with only the loss $\mathcal{L}_{rec} = ||f(x)-y||_2$. For the HDR reconstruction application, the Sparse Jacobian formulation took on average 1.92 times longer, whereas the transform invariance took 1.99 times longer. The latter is slightly slower since it requires running the transformation $T(f(x))$ on the reconstructed image $f(x)$.

\section*{S4: Experimental setup}
The two different applications used for the experiments are evaluated in the following way:
\begin{itemize}
	\item In the total loss in \eqnref{loss}, we use three different formulations of $\mathcal{L}_{reg}$: stability \eqref{eqn:stability_loss}, transform invariance \eqref{eqn:warp_loss}, and sparse Jacobian \eqref{eqn:jacobian_loss} regularization.
	\item The regularization strength is sampled at 12 locations, $\alpha_i = \frac{l_i}{l_i+1}$, $i=1,...,12$, where $l_i = 2^{i-3}$. This means that the relative regularization strength, or ratio $\frac{\mathcal{L}_{reg}}{\mathcal{L}_{rec}}$, will double for each point.
	\item For the perturbed sample $T(x)$, we use the geometric transformation specifying the warping from coordinate transformations according to \eqnref{transf}. For the stability regularization we also add one setting with noise perturbations, $T(x) = x + \Delta{x}$, where $\Delta{x} \sim \mathcal{N}(0,\sigma)$, and $\sigma$ is randomly selected for each image, $\sigma \sim \mathcal{U}(0.01,0.04)$.
	\item We complement with a training run using $T(x)$ for specifying na\"ive augmentation, increasing the training dataset size from $N$ to $2N$.
	\item For each combination of the above, we run 10 individual trainings, in order to estimate a proper mean and standard deviation of each datapoint.
\end{itemize}

In total, the combinations and repeated runs means that for each of the two applications we perform 500 optimization runs.

\end{document}